# TECHNIQUES FOR ADVERSARIAL EXAMPLES THREATENING THE SAFETY OF ARTIFICIAL INTELLIGENCE BASED SYSTEMS

Utku Kose

*Suleyman Demirel University, Faculty of Engineering, Dept. of Computer Engineering*

*utkukose@sdu.edu.tr*

**Abstract**

Artificial intelligence is known as the most effective technological field for rapid developments shaping the future of the world. Even today, it is possible to see intense use of intelligence systems in all fields of the life. Although advantages of the Artificial Intelligence are widely observed, there is also a dark side employing efforts to design hacking oriented techniques against Artificial Intelligence. Thanks to such techniques, it is possible to trick intelligent systems causing directed results for unsuccessful outputs. That is critical for also cyber wars of the future as it is predicted that the wars will be done unmanned, autonomous intelligent systems. Moving from the explanations, objective of this study is to provide information regarding adversarial examples threatening the Artificial Intelligence and focus on details of some techniques, which are used for creating adversarial examples. Adversarial examples are known as training data, which can trick a Machine Learning technique to learn incorrectly about the target problem and cause an unsuccessful or maliciously directed intelligent system at the end. The study enables the readers to learn enough about details of recent techniques for creating adversarial examples.

## 1. Introduction

In today's modern world, it is possible to see examples of Artificial Intelligence based applications in all fields of the life. Changing from engineering sciences to medical or education to entertainment, intelligent systems, which are developed with approaches, methods, and techniques of Artificial Intelligence, improves standards of our life and ensures more practical ways for solving real world problems [1-3]. Although there is a long past history behind the field of Artificial Intelligence, rise of computer, electronics, and communication technologies has triggered rapid improvements within the field of Artificial Intelligence [4, 5]. Nowadays, 21$^{st}$ century has already become a wide playground for rapid and innovative developments by Artificial Intelligence. It is remarkable that people generally benefits from Artificial Intelligence oriented tools with no doubt and thinks that a good future is possible thanks to intelligent systems (including robotic assistants, robotic workers, intelligent software systems…etc.). However, the scientific community has been running a great discussion for a while, regarding possible dangers of intelligent systems and precautions to eliminate such dangers. When the current state is analyzed, it can be seen that the scientific community has divided into two groups: the first group thinking that the Artificial Intelligence is dangerous, and the second group thinking otherwise. Even some scientists and famous leaders of some technology companies have enrolled in serious discussions with totally jarring opinions [6-8]. That's all because of flexible and robust learning capability of Artificial Intelligence. Today, there are many different sub-fields of Artificial Intelligence, such as Swarm Intelligence (for intelligent optimization) [9, 10], Cybernetics (for combining biological and robotic systems) [11, 12], and Machine Learning [13, 14]. Here, the sub-field of Machine Learning employs the learning mechanism, which is making it the major sub-field of the Artificial Intelligence.

When the research works including Machine Learning are observed, it can be clearly seen that the success of every Machine Learning technique is associated greatly with the data (samples) used for learning about the target problem. So far, successful applications by Machine Learning techniques have been done thanks to good modeling of the target problem and using true enough data. A typical Machine Learning technique employs some parameters, which are optimized-adjusted according to a set of target data, whether it was obtained in the past or analyzed instantly. Because of that, the background of the Machine Learning is highly associated with Mathematics, Statistics, and Logic (like all other Artificial Intelligence techniques, as the algorithm oriented perspective required that generally) [15-18]. It is critical that all that background gives both some advantages and disadvantages to the Machine Learning. Essential advantages are being flexible and easy-to-use for even different, advanced problems of real world. On the other hand, disadvantages include requiring careful analyze of the data, using sometimes pre-processing for the data, and also need for correct expert knowledge for true modeling of the problem. But as a result of revolutionary developments in data sharing, transfer and flow through communication roads around the world, security issues have raised against Machine Learning. As like threats in terms of cyber security, intelligent systems of today's and the future may employ some gaps making them weak against some attacks designed with strong mathematical and logical background (as similar to the background giving the expressed advantages). That means there is also a dark side employing efforts to design hacking oriented techniques against Artificial Intelligence and its sub-field: Machine Learning. Thanks to such techniques, it is possible to trick intelligent systems causing directed results for unsuccessful outputs. That is critical for also cyber wars of the future as it is predicted that the wars will be done unmanned, autonomous intelligent systems. When they are examined in terms of especially Machine Learning, some techniques focusing on creating fake learning-training data (samples), which may trick target intelligent system (employing Machine Learning mechanism) have already taken their place in the scientific literature. All these techniques are currently examined under the topic of 'adversarial examples'.

Objective of this study is to provide information regarding adversarial examples threatening the Artificial Intelligence and focus on details of some techniques, which are used for creating adversarial examples. Adversarial examples are known as training data, which can trick a Machine Learning technique to learn incorrectly about the target problem and cause an unsuccessful or maliciously directed intelligent system at the end. In this context, the study aimed enabling the readers to learn enough about details of recent techniques for creating adversarial examples. In order to achieve that, some essential information regarding Machine Learning and explanations for currently known, recent techniques for adversarial examples have been given accordingly.

Based on the topic of the study, the remaining content of the paper is organized as follows: The next section simply explains some about what is the general logic on the background of Machine Learning based techniques and systems. That section also includes some brief information regarding how different learning-training paradigms are seen mathematically. Following to that section, the third section discusses about Artificial Intelligence Safety and explains adversarial examples with some known recent techniques. Next, the fourth section provides a sample for a specific technique of creating adversarial examples and then the paper is ended by discussions regarding conclusions, and some future work suggestions-ideas.

## 2. Background of Machine Learning Based Systems

As it is already widely applied and discussed in real world cases greatly, Machine Learning has proved its active role in the future of intelligent systems. Whether they are in the form of software systems or real robots, intelligent systems having the mechanisms of autonomous analyzing, decision making and acting are based on Machine Learning techniques-algorithms. Because of that, such systems need to interact with the real world for analyzing some data to learn (or be trained) better about even a newly encountered problem. That interaction may be examined in detail, under different research fields such as Human-Computer or Human-Robot Interaction [19, 20] but it is possible to simply indicate that a typical Machine Learning based system is responsible to use data for learning, and getting that data instantly or with human controlled feeding, in order to solve problems even their conditions are changed, and eventually run a proper life cycle in that flow.

In the context of the happening interaction, there is a great flow of learning from data, which is the main task of all Machine Learning techniques. Although different techniques are based on different algorithmic steps and parameters to be optimized-adjusted, the common thing among all of them is to be a learned (trained) model eventually. That logic has been explained in a simpler form under the following sub-section.

### 2.1. The Logic of Learning

In order to understand better about how the 'learning' is achieved in Machine Learning techniques, the learning mechanism in humans may be given as an example. If we eliminate complex details such as learning theories or learning psychology, it is possible to express that the learning in humans is just happened as result of adjustments done within electrical interactions among all cells and the related organs in the nervous system. It is important that the brain has the responsibility of storing all information that we may need later to solve problems we may encounter in the future. It is critical that the more we focus on learning some information the more our brain can store about it for possible recalling in the future. Any information stored in the brain represented with a group of cells (in terms of thousands or millions of cells) having electrical connections among them. The strength level of these electrical connections correspond to the possibility of recalling-remembering the associated information anything we need. Moving from that, it is possible to indicate that:

- We remember good or bad memories because they affected our brain too much so that very high electrical connections allow us to remember them almost always.

- We may forget some details for even good or bad memories because our brain stores only important points affected us too much.

- When we are studying something, it is possible to memorize it if we give enough time to repeat all about it and allow our brain to employ necessary cells and create electrical connections for storing.

- Memorizing is effective in short terms but electrical connections formed during memorizing may be lost if we do not study again and again.

- It is possible to say that better learning about something is associated with encountering with more parameters-factors associated with it. Also, electrical connections are created in a stronger way, when we do or experience something.

All the explained things above may be somehow learned from medical, biology, chemistry or cognitive science oriented sources but the author have decided to express them in his own words, by eliminating scientific details.

Frankly speaking, the brain and also many components of the nature have always been essential elements of all technological developments. From data editing and storing capabilities of computer systems to learning mechanisms in the Machine Learning, inspirations from them have had critical roles on developments. As similar, the logic in the context of Machine Learning is mainly related to the learning process in the brain. In detail, it is possible to explain more about that logic by connecting it with the example expressed under the previous paragraphs:

- As the brain responsible for storing information, Machine Learning employs great storage power of computer systems, by running also data structures or models for effective coding in terms of software side.

- Machine Learning techniques are designed in the form of algorithms, which can adjust some of their parameters (variables) or order of data models iteratively (in loops). These parameters or data models correspond to the electrical connections among cells in the brain / nervous system.

- Every Machine Learning technique is designed in the form of a system analyzing some inputs to get an output (problem solution) or some outputs (more than one solution for the target problem). That is like we read, listen or watch something and then our brain can process it to derive some findings, ideas eventually.

- As how we study or do something to learn about it or solution of a problem, a typical Machine Learning technique is run iteratively by feeding it with a set of data associated with a problem. These data may be different cases for the related problem and each output data obtained after each different input feed are analyzed with some metrics.

- As like how we can understand from results of exams, fails (after our actions), and instant feedback from environment, output-analyzing metrics give some values to update parameters (variables) or order of the data models included in the Machine Learning technique. That update is just an optimization-adjustment, which is rechecked again to see if the technique is later better on giving correct output(s). That is like we study more and more for a course we just failed. In terms of Machine Learning, it is all called as 'training' or 'learning' phase.

- In the real world, humans pass schools, learn something from courses, and try to give some exams to prove that they learned something correctly. As similar, the training-learning process of Machine Learning techniques is done with some training data, and then some other sample data were used for testing if the technique was trained correctly. That is done because the mathematical and logical infrastructure of the Machine Learning may cause overfitting, which means memorizing the training data [21, 22]. In that case, the technique is not successful in test data because it has just memorized. It is again similar for humans' learning process as if we memorize something, it may be effective in short term but does not mean we will pass the main exam or will be always successful on solving the related problem. Thanks to some solutions, Machine Learning techniques can prevent from overfitting (like humans may take additional courses or be tested from different perspectives to promote better studying and eliminate memorizing). If a Machine Learning technique is successful in also testing phase, it can be then accepted to be applied in real applications. Eventually, it can be said that using cycle of all Machine Learning techniques includes the phases of 'training', 'testing', and 'application' (Figure 1-a).

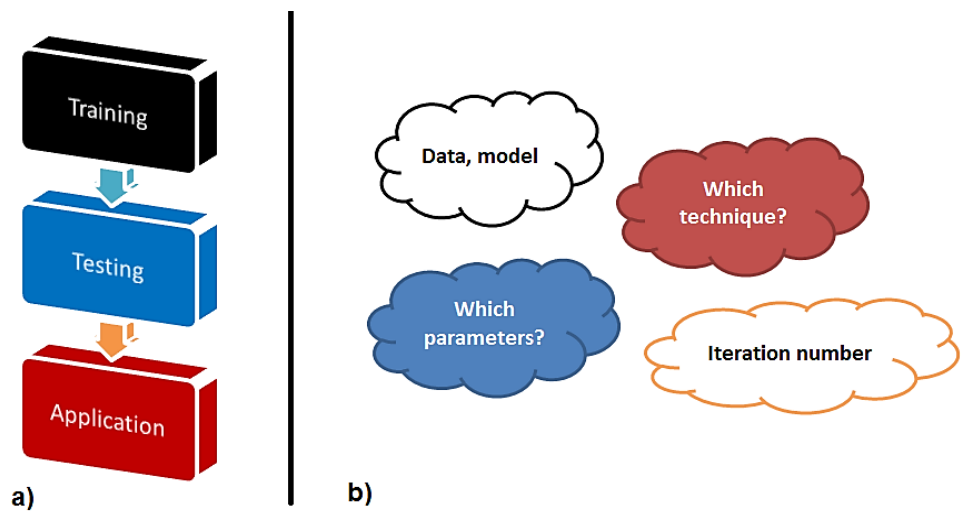

**Figure 26: a) Using cycle of Machine Learning techniques, b) Issues regarding using Machine Learning.**

As it was mentioned before, the data for learning and modeling the target problem are some essential issues that should be considered in research works with Machine Learning. Except from them, it is also important to choose the most appropriate Machine Learning technique since some techniques may not be effective on some specific problems. Furthermore, since Machine Learning techniques may employ their own additional parameters (except from the optimized ones); it is another issue to choose the most appropriate value for such parameters. Finally, total iteration number for better learning or the need for running a pre-processing phase for the data are some other issues that should be considered (Figure 1-b).

All the mentioned logic of the learning mechanism (within Machine Learning) are designed by some mathematical and logical background. Here, it is also remarkable that the associated literature have some well-known learning-training paradigms, which are run according to structure of the target data during training phases. The next sub-section is devoted to these paradigms.

**2.2. Learning-Training Paradigms**

Because every data-information gathered from real world may not be always in same structure, Machine Learning uses different learning (training) paradigms while making them ready to be applied. In this sense, there is a relation between Artificial Intelligence and Data Mining since roots of learning paradigms are associated with similarities among data, their distributions and patterns that can be derived from them [23, 24]. Currently, there are three main learning paradigms, which can be explained as follows:

### 2.2.1. Supervised Learning (Regression / Classification)

Supervised learning is the process of learning when outputs for each different input data are known. In other words, each known input case for the considered problem is labeled with known target output class(es). In Data Mining, that is called as also regression and / or classification, which means we can predict a value by figuring out relations among known values or classify results of different input (attribute) combinations, considering the target problem [25, 26]. That means a successfully trained Machine Learning technique with supervised learning can predict a value or make classification according to a newly encountered input combination. In mathematical terms, a regression model of prediction can be obtained for deriving value(s) based on relations among samples or a two-class problem with linear distribution can be learned by a linear equation indicating the border between the two classes. In classification, it can be also necessary to deal with more than two classes with even nonlinear distribution of the known samples (data), which requires some mathematical processes in higher dimensions. Figure 2 represents typical a) regression and b) nonlinear classification in terms of mathematical and visual view, thanks to supervised learning [27, 28].

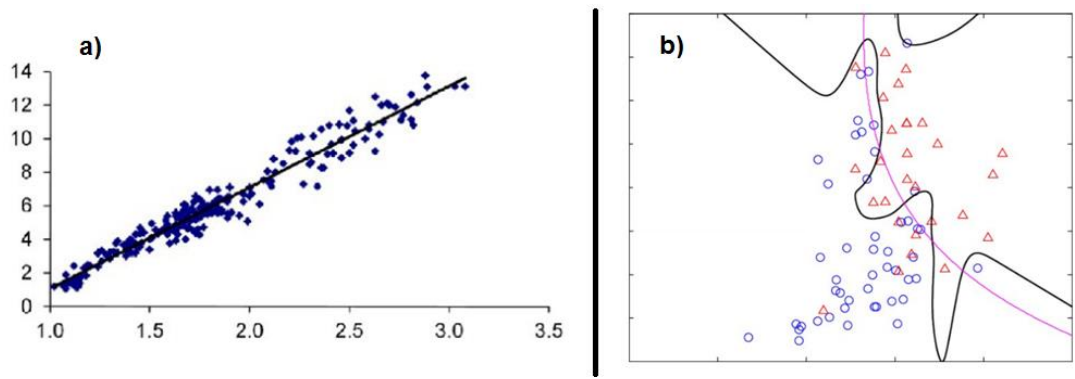

**Figure 27: a) Regression, b) Nonlinear classification with supervised learning [27, 28].**

### 2.2.2. Unsupervised Learning (Clustering)

Unsupervised learning is done when outputs for the target data to be learned from are not known. In this case, similarities-distances among the data are used for grouping them into some determined number of clusters. Because of that, the term of clustering is used by Data Mining for defining that way of learning from data. Unsupervised learning (clustering) is like we know only final grades by some students and want to group them into i.e. five groups, by considering distances between the values and also determining some passing values between each two different groups. Mathematically speaking, similarities-distances among the data are determined with some known metrics such as Manhattan, Minkowski, and Euclidean Distance [25, 26]. Unsupervised learning here allows determining the cluster-group of a new input data by evaluating its similarity-distance according to the formed groups during training phase. Figure 3 provides a typical clustering approach applied over a set of data [29].

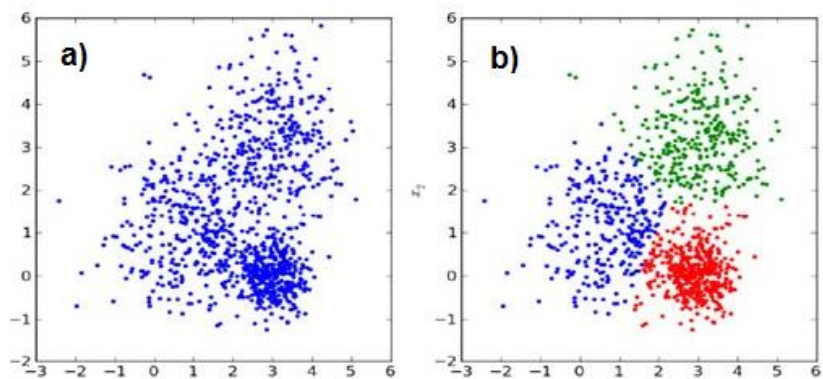

**Figure 28: a) Before, b) After an unsupervised learning (clustering) [29].**

### 2.2.3. Reinforcement Learning

Reinforcement learning is known as a key learning paradigm for especially robotic systems of the future. In that learning paradigm, the Machine Learning based system has the responsibility to adapt some of its parameters according to the given feedback against results of its actions. So, it is possible to run the learning process instantly, according to the changing conditions of the environment [30, 31]. Because each feedback given to the intelligent system is determined according to the strategy of rewarding or punishing ('true / false'; 'correct' / 'incorrect'; 'good', 'normal', 'bad'…etc.), it is important to give accurate feedback for better evolved system. Mathematically speaking, that learning process includes calculating values of reward in case of performing some target actions, and / or organizing a scheme or data model of actions for future considerations. Since that learning paradigm is associated with also agent based systems (which we may think as early action models of advanced robotic systems), Figure 4 represents a general flow of active reinforcement learning [32].

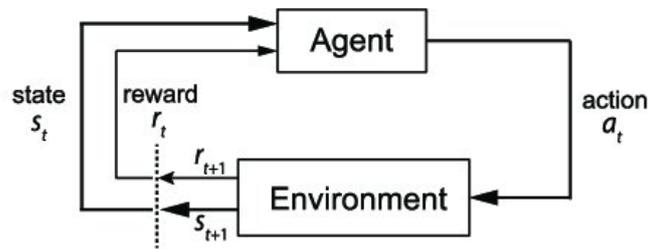

**Figure 29: General flow of active reinforcement learning [32].**

Except from the expressed paradigms, there is also Semi Supervised Learning, which is a combination of both Supervised and Unsupervised Learning as including more data with no output-label, and less data with output(s)-label(s) for the same problem [33]. But three essential paradigms remain main learning processes of Machine Learning.

As it can be understood from the explanations so far, there is a great Mathematics and Logic in terms of everything used for Machine Learning tasks. Because of that, there can be some gaps causing to manipulate intelligent systems with Machine Learning.

**2.3. Gaps in Machine Learning Based Systems**

It is clear that the Artificial Intelligence and Machine Learning are all human made technologies so they may include human errors but be always open for improvements thanks to human touches. On the other hand, we can discuss briefly about some gaps caused by the exact nature of algorithms and the background of the Machine Learning. In this context, some gaps are:

- For advanced and critical problems, Machine Learning based systems should be often updated-upgraded in terms of designed problem model and used data.

- Machine Learning techniques are very sensitive to butterfly effects caused by small errors in training / test data.

- Especially Reinforcement Learning has the potential of training 'bad' intelligent systems when we theoretically speak. That is like bad people, who were actually baby sometimes ago but grown up badly in time because of bad environmental factors (bad people, low life standards) and especially lack of good enough education and psychological support. For Reinforcement Learning, there are some works for preventing intelligent systems from badly trained (i.e. Inverse Reinforcement Learning [34]) but that is still a gap if an intelligent system with Machine Learning start to receive incorrect or tricking feedback after a while.

- In terms of solving real world problems, not every Machine Learning technique is equally successful. That's good for unstoppable research works and future diversities within the literature of Artificial Intelligence but that also shows there is not any universal mathematical infrastructure to design a Machine Learning technique (or maybe learning paradigm), which can be applied directly every kind of problem without doubt. Actually, that situation may be a sign for open problems in Mathematics, Statistics, Logic, or other supportive components of the Artificial Intelligence.

- That may be a success for Artificial Intelligence and Machine Learning but especially after rise of Deep Learning (advanced form of Machine Learning with more power against big, complex data [35]), it has become possible to develop intelligent systems, which can create very realistic photos, and even videos with montage (i.e. Deep Fake [36]). At this point, there is a critical question to ask: 'Is it possible to have a good enough intelligent system, which can overcome dangerous applications such as Deep Fake?'. Whether the answer is 'Yes' or 'No', that's a paradox, which we may accept as a gap that can be used against Machine Learning.

- Flexibility and high interaction of Machine Learning techniques with the data make them sensitive to fake data formations or direct / indirect use of mathematical and logical attacks to trick them.

Especially the last gap indicated has already found its way in the context of adversarial examples, which is used for defining dangerous, fake data (samples) to train Machine Learning techniques incorrectly. Moving from that, the next section explains the concept of Artificial Intelligence Safety and some known techniques for creating adversarial examples.

**3. Artificial Intelligence Safety and Techniques for Adversarial Examples**

As a result of ongoing discussions regarding future potential of Artificial Intelligence (in being safe or dangerous), some research fields have already taken active place in the scientific literature. Because the field of Artificial Intelligence has become also a multidisciplinary technology (covering both Natural and Social Sciences), it has been included in some general research fields followed by the scientific community for a long time. Among all these research fields, **Artificial Intelligence Safety** is enrolled in searching for possible dangerous sides of Artificial Intelligence based systems and introduce alternative approaches, methods, and techniques to keep such systems from attacks, perform defensive attacks, or eliminate known gaps against hacking or manipulating oriented attacks [37-40]. For giving a wide enough theoretical background, some other research fields are as follows:

- **Machine Ethics / Ethical Artificial Intelligence:** The concept of ethics has been widely discussed in the context of Artificial Intelligence literature. Because of intense intersection with different fields such as philosophy, and psychology, research efforts on evaluating ethical applications of Artificial Intelligence have been running in the context of the research field: Machine Ethics. Moving from especially moral dilemmas, that field tries to find ways to solve paradoxes and design ethical applications of Artificial Intelligence, particularly Machine Learning [41, 42].

- **Superintelligence:** As firstly introduced by Prof. Nick Bostrom from Future of Humanity Institute (University of Oxford), the concept of Superintelligence is used for defining the most advanced and high-level form of Artificial Intelligence that is more intelligent than the most intelligent living organism over the world (or even universe) [43, 44]. As Superintelligence is theoretically strong enough to solve every kind of problem and greatly autonomous, possible dangers and benefits of that type of Artificial Intelligence have been discussing accordingly.

- **Technological Singularity:** As a remarkable hypothesis, Technological Singularity is associated with the radically changed form of society and the world in the future. According to the Technological Singularity, the future will be dominated by Superintelligence and that will transform the society, technological developments, and the world-universe to a different form, which is thought as utopia or dystopia by the community [45, 46].

- **Future of Artificial Intelligence:** As a wider research field, Future of Artificial Intelligence is based on predictive and theoretical works trying to keep a light over the future with intelligence systems [47, 48].

- **Existential Risks:** Existential Risks is actually a research field in which disasters, wars, atomic and chemical weapons-bombs are widely examined as existential risks for the humankind. As a result of rising anxieties regarding dangerous sides of intelligent systems, Artificial Intelligence has been added to the list of existential risks [49, 50].

- **Transhumanism:** Transhumanism is known as both wide-spread movement and research field focusing on a better future for the humankind. In that future idea, there is a great use of technology so that humans' life standards, well-being and cognitive levels are enhanced accordingly [51, 52].

As different from the mentioned research fields, Artificial Intelligence Safety is a more complicated field with intense works more on mathematical background and applications of especially Machine Learning. In this study, more consideration was given to the concept of adversarial examples. So, the next sub-fields are devoted to techniques of adversarial examples. On the other hand, interested readers are referred to also some alternative Artificial Intelligence Safety oriented, remarkable works such as 'Interruptible Agents [53]', 'Ignorant, Inconsistent Agents [54]', 'Bounded Agents [55]', and 'Corrigibility [56]'.

### 3.1. Techniques to Create Adversarial Examples

As an attacking-hacking method for tricking Machine Learning techniques, the concept of adversarial examples was explained firstly by Szegedy et al. [57]. As indicated in the related work, it has been possible to trick Deep Neural Network models by making small changes in input data. After that, there has been a great momentum in terms of developing defensing techniques and attacking techniques defeating every new defensing technique, by considering adversarial examples [58]. As it can be seen from Figure 5, total number of new attacking techniques increases year-by-year.

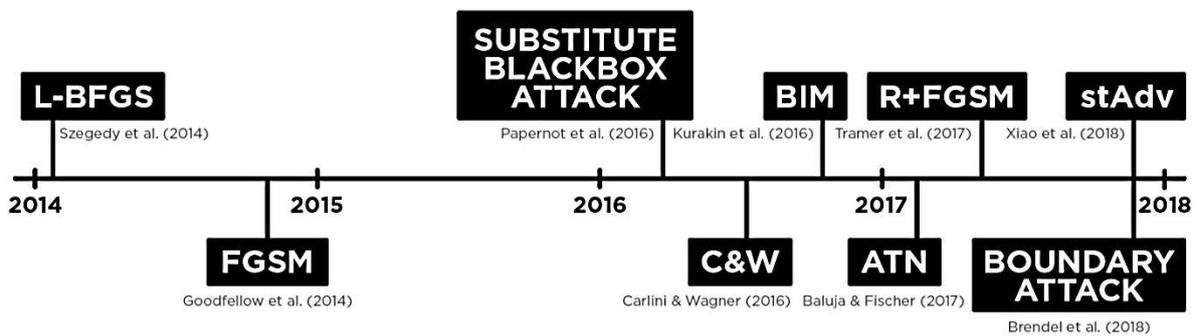

**Figure 30: A historical view regarding new techniques of adversarial examples [58].**

In the associated literature, there are different examples of using adversarial examples threating different applications by intelligent systems. The research work by Sharif et al. showed how adversarial examples can trick face recognition systems [59]. On the other hand, the work by Carlini and Wagner figured out that speech recognition systems can be tricked with audio adversarial examples [60]. Additionally, another remarkable work included using a real traffic sign and manipulating it with some stickers to trick autonomous vehicles successfully [61]. Readers are referred to [62-68] for some other works including use of adversarial examples.

Generally, techniques for adversarial examples can be examined under two categories [58]:

- **White-box Techniques:** White-box techniques are appropriate to be used when the architecture and/or parameters of the target Machine Learning technique are known.
- **Black-box Techniques:** Black-box techniques are appropriate to be used when there is no information regarding architecture or parameters of the target Machine Learning technique but output(s) of the technique are known.

Considering the recent literature, some of **White-box techniques** can be explained briefly as follows:

*3.1.1. Additive Adversarial Perturbations Based on dL/dx*

In this technique, it is aimed to confuse the input(s) so that they the loss function is changed maximally. In other words, it is possible to run a backpropagation oriented approach, which includes calculating the derivative of the loss function according to the input(s) [58]. That's actually an optimization approach, which is done for training the target Machine Learning technique in a way it learns incorrectly, as maximizing the loss.

*3.1.2. Fast Gradient Sign Method (FGSM) and Other Variations*

As introduced by Goodfellow et al. [58, 69], Fast Gradient Sign Method (FGSM) is used for pushing the input(s) towards the adversarial gradient, after the exact gradient was calculated in the context of the backpropagation process. That is done in order to increase the loss function, by using a small enough value. The following equation defines the calculation of an adversarial example ($x'$) from the true value: $x$:

$$x' = x + \epsilon \cdot \text{sign}(\nabla_x J(x, y)) \tag{1}$$

As it can be seen from the Equation 1, the small epsilon parameter has the key role in creating the adversarial example, which is similar enough to the true value-example (input).

Moving from the FGSM, some other variations have also been developed. In this sense, Kurakin et al. developed the **Basic Iterative Method (BIM)**, by improving the FGSM for performing it multiple times with small step size [58, 70]:

$$X_0^{adv} = X, \quad X_{N+1}^{adv} = Clip_{X,\epsilon}\left\{X_N^{adv} + \alpha \, \text{sign}(\nabla_X J(X_N^{adv}, y_{true}))\right\} \tag{2}$$

There is also the technique of **(R)andom + FGSM**, which includes adding random perturbations derived from a Gaussian distribution just before obtaining the first derivative of the loss [58, 71]:

$$x^{adv} = x' + (\varepsilon - \alpha) \cdot \text{sign}\left(\nabla_{x'} J(x', y_{true})\right), \quad \text{where} \quad x' = x + \alpha \cdot \text{sign}(\mathcal{N}(\mathbf{0}^d, \mathbf{I}^d)) \tag{3}$$

*3.1.3. L-BFGS Attack and Carlini & Wagner Attack (C&W)*

By considering distance metrics, Szegedy et al. have introduced that adversarial examples, which are similar to true values-inputs can be created accordingly [57]. As in the form of an optimization problem, their technique called as Limited-memory Broyden-Fletcher-Goldfarb-Shanno (L-BFGS) is just a non-linear gradient based optimization algorithm. Here, the objective is to find the perturbation $r$ minimizing the following [57, 58]:

$$c|r| + \text{loss}_f(x + r, l) \text{ subject to } x + r \in [0, 1]^m \tag{4}$$

By following the [57], Carlini and Wagner have modified the optimization problem, in order to develop the technique of Carlini & Wagner Attack (C&W), which considers the loss function in Equation (5) and the minimization problem expressed with Equation (6) [58, 72]:

$$f(x') = \max(\max\{Z(x')_i : i \neq t\} - Z(x')_t, -\kappa) \tag{5}$$

$$\left\|\frac{1}{2}(\tanh(w) + 1) - x\right\|_2^2 + c \cdot f\left(\frac{1}{2}(\tanh(w) + 1)\right) \tag{6}$$

*3.1.4. Adversarial Transformation Network (ATN)*

As an interesting technique of adversarial examples, Adversarial Transformation Network (ATN) is a type of neural network, which can create adversarial examples tricking the target neural network system [73]. As a type of ATN, Adversarial Autoencoding creates adversarial examples, which are very similar to the true values-inputs (Figure 6) [58].

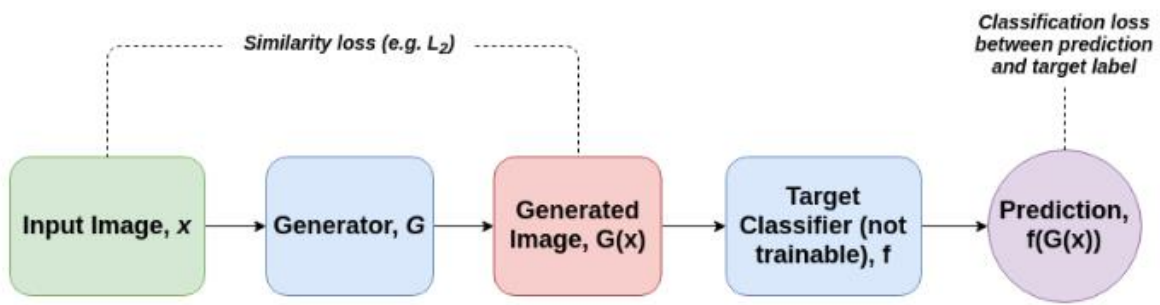

**Figure 31: Based on an image data, general working mechanism of the Adversarial Autoencoding [58].**

In the context of Black-box techniques, two essential techniques of adversarial examples are as follows:

*3.1.5. Substitute Attack*

When just output(s) of the target Machine Learning system are known, it is possible to create some synthetic training data sets in order to approximate decision boundary. When that is done successfully, it is possible to run White-box attacks for tricking the target system with newly created adversarial examples. As examined by Papernot et al., that Black-box technique is called as Substitute Attack [74]. Figure 7 shows the simple flow of the Substitute Attack [58].

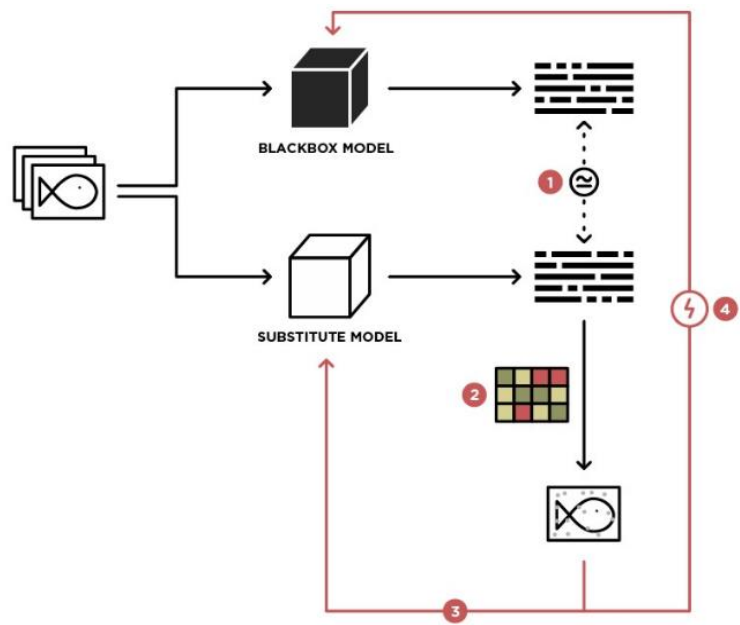

**Figure 32: Simple flow of the Substitute Attack [58].**

*3.1.6. Heuristic Search for Boundary Attack*

Boundary attack with heuristic search is another technique of Black-box oriented adversarial examples. Called briefly as Boundary Attack [64], that technique aims to use a fake data (i.e. image) and make some changes over it to look similar to true data for the target Machine Learning technique. Thanks to some noisy data, that can be achieved and that adversarial example then can be used for tricking purposes. Figure 8 represents a scheme regarding applying the Boundary Attack [58].

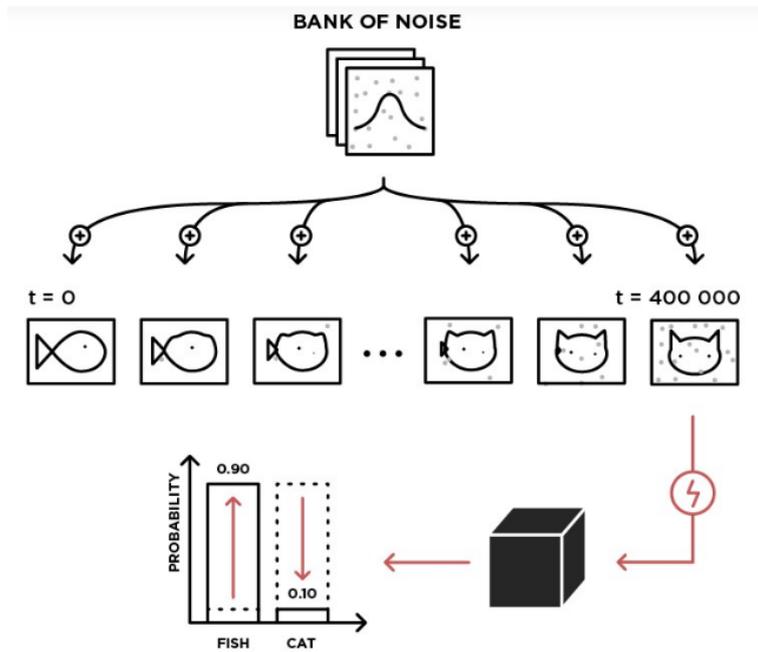

**Figure 33: A scheme regarding applying the Boundary Attack [58].**

**4. An Example Application for Fast Gradient Sign Method (FGSM)**

As a simple way of applying adversarial examples, FGSM is a good and easy-to-apply technique. Although there are many examples of codes and applications provided over the Web for learning some about adversarial examples, a simple Python application of FGSM, done by Tsui may give idea about how adversarial examples are created [75]. As it was indicated before, FGSM just runs the Equation (1) for creating adversarial values from true values. The Python application in [75] tricks the Logistic Regression as the target Machine Learning technique (Full source code can be obtained from the developer's GitHub folder: https://github.com/kenhktsui/adversarial_examples). As it can be seen from Figure 9, the true data was classified into two classes, by the Logistic Regression (in the example, random data points are created as the problem data set) [75].

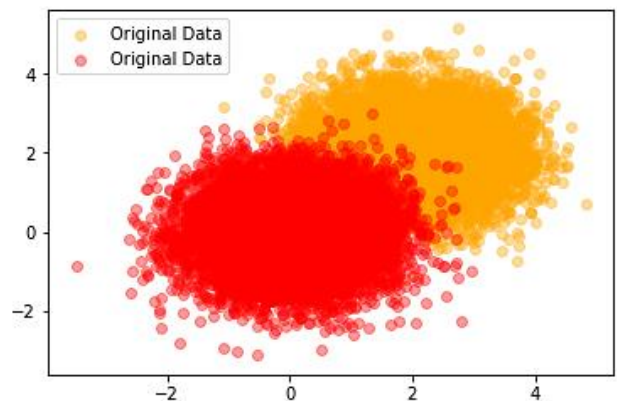

**Figure 34: The true data classified into two classes, by the Logistic Regression (As using the application by Tsui in [75]).**

By using the Equation (1), it is possible to create adversarial examples causing the target Logistic Regression system to classify the data points incorrectly. In order to achieve that, there is need to adjust the epsilon parameter, which is causing small but effective enough changes over the data, as making them similar to the true-original data but tricking the system. By moving in the example application, use of even the values of 0.4, and 0.7 respectively causes great changes in the output (like a butterfly effect). In this context, Figure 10 presents the outputs in the true classification and incorrect classifications for two different epsilon values, considering the randomly created data points.

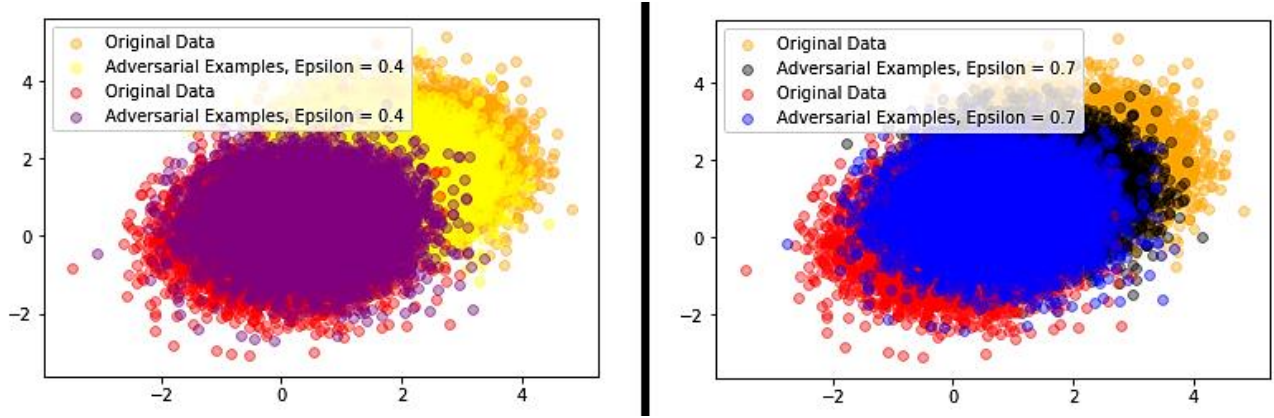

**Figure 35: Outputs in the true classification and incorrect classifications for two different epsilon values of 0.4, and 0.7 respectively (As using the application by Tsui in [75]).**

**5. Conclusions and Future Work Suggestions**

In this study, techniques for adversarial examples threatening safety of especially Machine Learning based systems have been introduced. Adversarial examples are typical form of manipulations, which can trick Machine Learning techniques to give incorrect outputs for the target problem. The study briefly introduced the logical background of the Machine Learning and then given brief information about the field of Artificial Intelligence Safety and some other fields interested in similar topics regarding Artificial Intelligence. After explaining both White Box and Black Box techniques for creating adversarial examples, an example from the Web for the technique of Fast Gradient Sign Method (FGSM) has been discussed accordingly. Although the works regarding Artificial Intelligence Safety requires intense analyze and use of especially Mathematics, a remarkable movement has been done for the topic of adversarial examples. Of course, it is a just 5-year old research topic so far and there is currently still great focus on running alternative research works.

By moving from the discussed-explained subjects within this study, the following future work suggestions may be expressed:

- As it can be seen from recent works in the literature, there is a great competition between attacking and defensing techniques. So, interested readers-researchers are suggested to develop alternative techniques in terms of both attacking and defensing, in order to contribute to the rapid development of the field of Artificial Intelligence and the research field of Artificial Intelligence Safety.

- It is necessary to form collaborations among computer scientists, mathematicians, statisticians, and any experts from even Social Sciences, in order to derive novel, effective theoretical and applied techniques for keeping the safety of intelligent systems against attacks such as adversarial examples.

- Research works regarding adversarial examples and defensing approaches are mainly focused on neural networks of Machine Learning. So, alternative attacking techniques for different Machine Learning techniques may be developed in time. That will be probably an open problem for analyzing such new attacking techniques and even designing robust systems or defensing ways to eliminate the related attacks.

- Artificial Intelligence Safety has many other topics to be discussed in the context of ensuring safety of intelligent systems. At this point, interested readers-researchers are suggested to go through the associated literature and learn more about that research field.

**6. References**

[1] Rabelo, L., Bhide, S., and Gutierrez, E., "*Artificial Intelligence: Advances in Research and Applications*". Nova Science Publishers, Inc., 2018.

[2] Russell, S. J., and Norvig, P., "*Artificial Intelligence: A Modern Approach*". Malaysia; Pearson Education Limited, 2016.

[3] Barr, A., and Feigenbaum, E. A. (Eds.), "*The Handbook of Artificial Intelligence*" (Vol. 2). Butterworth-Heinemann, 2014.

[4] Nilsson, N. J., "*Principles of Artificial Intelligence*". Morgan Kaufmann, 2014.

[5] Krishnamoorthy, C. S., and Rajeev, S., "*Artificial Intelligence and Expert Systems for Engineers*. CRC Press, 2018.

[6] Cellan-Jones, R., "Stephen Hawking warns artificial intelligence could end mankind". *BBC News*, *2*, 2014.

[7] Walsh, T., "*Android Dreams: The Past, Present and Future of Artificial Intelligence*". Oxford University Press, 2017.

[8] Delnevo, G., Roccetti, M., and Mirri, S., "Intelligent machines for good?: More focus on the context". In *Proceedings of the 4th EAI International Conference on Smart Objects and Technologies for Social Good* (pp. 289-293). ACM, 2018.

[9] Bonabeau, E., Marco, D. D. R. D. F., Dorigo, M., and Theraulaz, G., "*Swarm Intelligence: From Natural to Artificial Systems*. Oxford University Press, 1999.

[10] Eberhart, R. C., Shi, Y., and Kennedy, J., "*Swarm Intelligence*". Elsevier, 2001.

[11] Novikov, D. A., "*Cybernetics: From Past to Future*" (Vol. 47). Springer, 2015.

[12] Kline, R. R., "*The Cybernetics Moment: Or Why We Call Our Age the Information Age*. JHU Press, 2015.

[13] Sammut, C., and Webb, G. I., "*Encyclopedia of Machine Learning and Data Mining*". Springer Publishing Company, Incorporated, 2017.

[14] Flach, P., "*Machine Learning: The Art and Science of Algorithms that Make Sense of Data"*. Cambridge University Press, 2012.

[15] Alpaydin, E., "*Introduction to Machine Learning*". MIT Press, 2009.

[16] Cheeseman, P., and Oldford, R. W. (Eds.), "*Selecting Models From Data: Artificial Intelligence and Statistics IV*" (Vol. 89). Springer Science and Business Media, 2012.

[17] Fisher, D., and Lenz, H. J. (Eds.), "*Learning from Data: Artificial Intelligence and Statistics V*" (Vol. 112). Springer Science and Business Media, 2012.

[18] Rahwan, I., and Simari, G. R. (Eds.), "*Argumentation in Artificial Intelligence*" (Vol. 47). Heidelberg: Springer, 2009.

[19] Preece, J., Rogers, Y., Sharp, H., Benyon, D., Holland, S., and Carey, T., "*Human-Computer Interaction*". Addison-Wesley Longman Ltd., 1994.

[20] Dautenhahn, K., and Saunders, J. (Eds.), "*New Frontiers in Human Robot Interaction*". John Benjamins Publishing, 2011.

[21] Dietterich, T., "Overfitting and undercomputing in machine learning". *ACM Computing Surveys*, *27*(3), 326-327, 1995.

[22] Lawrence, S., and Giles, C. L., "Overfitting and neural networks: conjugate gradient and backpropagation". In *Proceedings of the IEEE-INNS-ENNS International Joint Conference on Neural Networks. IJCNN 2000. Neural Computing: New Challenges and Perspectives for the New Millennium* (Vol. 1, pp. 114-119). IEEE, 2000.

[23] Tan, P. N., "*Introduction to Data Mining*". Pearson Education India, 2018.

[24] Witten, I. H., Frank, E., Hall, M. A., and Pal, C. J., "*Data Mining: Practical Machine Learning Tools and Techniques*". Morgan Kaufmann, 2016.

[25] Halgamuge, S. K., and Wang, L. (Eds.), "*Classification and Clustering for Knowledge Discovery*" (Vol. 4). Springer Science and Business Media, 2005.

[26] Silahtaroglu, G., "*Data Mining*" (In Turkish). Papatya Press, 2008.

[27] Wicks, J., McKenna, K., McSorley, S., and Craig, D., "Heart Rate Index Corrects for The Limitations of Heart Rate Assessment of Occupational Physical Activity". *Exercise Medicine*, *2*, 14, 2018.

[28] Yang, S., Cai, S., Zheng, F., Wu, Y., Liu, K., Wu, M., ... and Chen, J., "Representation of fluctuation features in pathological knee joint vibroarthrographic signals using kernel density modeling method". *Medical engineering and physics*, *36*(10), 1305-1311, 2014.

[29] Gattal, A., Abbas, F., and Laouar, M. R., "Automatic Parameter Tuning of K-Means Algorithm for Document Binarization". In *Proceedings of the 7th International Conference on Software Engineering and New Technologies* (p. 24). ACM, 2018.

[30] Kaelbling, L. P., Littman, M. L., and Moore, A. W., "Reinforcement learning: A survey". *Journal of Artificial Intelligence Research*, *4*, 237-285, 1996.

[31] Sutton, R. S., and Barto, A. G., "*Introduction to Reinforcement Learning*" (Vol. 2, No. 4). Cambridge: MIT Press, 1998.

[32] Galatzer-Levy, I. R., Ruggles, K. V., and Chen, Z., "Data science in the Research Domain Criteria era: relevance of machine learning to the study of stress pathology, recovery, and resilience". *Chronic Stress*, *2*, 2470547017747553, 2018.

[33] Zhu, X., and Goldberg, A. B., "Introduction to semi-supervised learning". *Synthesis lectures on artificial intelligence and machine learning*, *3*(1), 1-130, 2009.

[34] Abbeel, P., and Ng, A. Y., "Apprenticeship learning via inverse reinforcement learning". In *Proceedings of the twenty-first international conference on Machine learning* (p. 1). ACM, 2004.

[35] Goodfellow, I., Bengio, Y., and Courville, A., "*Deep Learning*". MIT Press, 2016.

[36] Güera, D., and Delp, E. J., "Deepfake video detection using recurrent neural networks". In *2018 15th IEEE International Conference on Advanced Video and Signal Based Surveillance (AVSS)* (pp. 1-6). IEEE, 2018.

[37] Yampolskiy, R. V., "*Artificial Intelligence Safety and Security*". Chapman and Hall/CRC, 2018.

[38] Yampolskiy, R. V., "Taxonomy of pathways to dangerous artificial intelligence". In *Workshops at the Thirtieth AAAI Conference on Artificial Intelligence,* 2016.

[39] Vassev, E., "Safe artificial intelligence and formal methods". In *International Symposium on Leveraging Applications of Formal Methods* (pp. 704-713). Springer, Cham, 2016.

[40] Köse, U., "Are we safe enough in the future of artificial intelligence? A discussion on machine ethics and artificial intelligence safety". *BRAIN. Broad Research in Artificial Intelligence and Neuroscience*, *9*(2), 184-197, 2018.

[41] Anderson, M., and Anderson, S. L. (Eds.), "*Machine Ethics*". Cambridge University Press, 2011.

[42] Moor, J. H., "The nature, importance, and difficulty of machine ethics". *IEEE Intelligent Systems*, *21*(4), 18-21, 2006.

[43] Bostrom, N., "*Super Intelligence: Paths, Dangers, Strategies*". Oxford University Press, 2014.

[44] Yampolskiy, R. V., "*Artificial Superintelligence: A Futuristic Approach*". Chapman and Hall/CRC, 2015.

[45] Kurzweil, R., "*The Singularity is Near: When Humans Transcend Biology*". Penguin, 2005.

[46] Shanahan, M., "*The Technological Singularity*". MIT Press, 2015.

[47] Müller, V. C., and Bostrom, N., "Future progress in artificial intelligence: A survey of expert opinion". In *Fundamental issues of artificial intelligence* (pp. 555-572). Springer, Cham, 2016.

[48] Walsh, T., "*Machines that Think: The Future of Artificial Intelligence*". Prometheus Books, 2018.

[49] Torres, P., and Rees, M., "*Morality, Foresight, and Human Flourishing: An Introduction to Existential Risks*". Pitchstone Publishing (US&CA), 2017.

[50] Müller, V. C. (Ed.), "*Risks of Artificial Intelligence*". CRC Press, 2016.

[51] Hansell, G. R., "*H+/-: Transhumanism and its Critics*". Xlibris Corporation, 2011.

[52] Huxley, J., "Transhumanism". *Ethics in Progress*, *6*(1), 12-16, 2015.

[53] Orseau, L., and Armstrong, S., "Safely interruptible agents". In *Proceedings of the ThirtySecond Conference on Uncertainty in Artificial Intelligence* (pp. 557-566). AUAI Press, 2016.

[54] Evans, O., Stuhlmüller, A., and Goodman, N., "Learning the preferences of ignorant, inconsistent agents". In *Thirtieth AAAI Conference on Artificial Intelligence,* 2016.

[55] Evans, O., and Goodman, N. D., "Learning the preferences of bounded agents". In *NIPS Workshop on Bounded Optimality* (Vol. 6), 2015.

[56] Soares, N., Fallenstein, B., Armstrong, S., and Yudkowsky, E., "Corrigibility". In *Workshops at the Twenty-Ninth AAAI Conference on Artificial Intelligence,* 2015.

[57] Szegedy, C., Zaremba, W., Sutskever, I., Bruna, J., Erhan, D., Goodfellow, I., and Fergus, R., "Intriguing properties of neural networks". *arXiv preprint arXiv:1312.6199,* 2013.

[58] Wiyatno, R. R., "Tricking a Machine into Thinking You're Milla Jovovich". Medium.com, 2018. Online: https://medium.com/element-ai-research-lab/tricking-a-machine-into-thinking-youre-milla-jovovich-b19bf322d55c (Retrieved 03 September 2019).

[59] Sharif, M., Bhagavatula, S., Bauer, L., and Reiter, M. K., "Accessorize to a crime: Real and stealthy attacks on state-of-the-art face recognition". In *Proceedings of the 2016 ACM SIGSAC Conference on Computer and Communications Security* (pp. 1528-1540). ACM, 2016.

[60] Carlini, N., and Wagner, D., "Audio adversarial examples: Targeted attacks on speech-to-text". In *2018 IEEE Security and Privacy Workshops (SPW)* (pp. 1-7). IEEE, 2018.

[61] Eykholt, K., Evtimov, I., Fernandes, E., Li, B., Rahmati, A., Xiao, C., ... and Song, D., "Robust physical-world attacks on deep learning models". *arXiv preprint arXiv:1707.08945,* 2017.

[62] Akhtar, N., and Mian, A., "Threat of adversarial attacks on deep learning in computer vision: A survey". *IEEE Access*, *6*, 14410-14430, 2018.

[63] Ilyas, A., Engstrom, L., Athalye, A., and Lin, J., "Black-box adversarial attacks with limited queries and information". *arXiv preprint arXiv:1804.08598,* 2018.

[64] Brendel, W., Rauber, J., and Bethge, M., "Decision-based adversarial attacks: Reliable attacks against black-box machine learning models". *arXiv preprint arXiv:1712.04248,* 2017.

[65] Lin, Y. C., Hong, Z. W., Liao, Y. H., Shih, M. L., Liu, M. Y., and Sun, M., "Tactics of adversarial attack on deep reinforcement learning agents". *arXiv preprint arXiv:1703.06748,* 2017.

[66] Narodytska, N., and Kasiviswanathan, S., "Simple black-box adversarial attacks on deep neural networks". In *2017 IEEE Conference on Computer Vision and Pattern Recognition Workshops (CVPRW)* (pp. 1310-1318). IEEE, 2017.

[67] Liu, Y., Chen, X., Liu, C., and Song, D., "Delving into transferable adversarial examples and black-box attacks". *arXiv preprint arXiv:1611.02770,* 2016.

[68] Finlayson, S. G., Chung, H. W., Kohane, I. S., and Beam, A. L., "Adversarial attacks against medical deep learning systems". *arXiv preprint arXiv:1804.05296,* 2018.

[69] Goodfellow, I. J., Shlens, J., and Szegedy, C., "Explaining and harnessing adversarial examples". *arXiv preprint arXiv:1412.6572,* 2014.

[70] Kurakin, A., Goodfellow, I., and Bengio, S., "Adversarial examples in the physical world". *arXiv preprint arXiv:1607.02533,* 2016.

[71] Tramèr, F., Kurakin, A., Papernot, N., Goodfellow, I., Boneh, D., and McDaniel, P., "Ensemble adversarial training: Attacks and defenses". *arXiv preprint arXiv:1705.07204,* 2017.

[72] Carlini, N., and Wagner, D., "Towards evaluating the robustness of neural networks". In *2017 IEEE Symposium on Security and Privacy (SP)* (pp. 39-57). IEEE, 2017.

[73] Baluja, S., and Fischer, I., "Adversarial transformation networks: Learning to generate adversarial examples". *arXiv preprint arXiv:1703.09387,* 2017.

[74] Papernot, N., McDaniel, P., Goodfellow, I., Jha, S., Celik, Z. B., and Swami, A., "Practical black-box attacks against machine learning". In *Proceedings of the 2017 ACM on Asia conference on computer and communications security* (pp. 506-519). ACM, 2017.

[75] Tsui, K., "Perhaps the Simplest Introduction of Adversarial Examples Ever". TowardsDataScience.com, 2018. Online: https://towardsdatascience.com/perhaps-the-simplest-introduction-of-adversarial-examples-ever-c0839a759b8d (Retrieved 04 September 2019).